\documentclass{article}
\usepackage{ijcai23}

\usepackage{times}
\usepackage{soul}
\usepackage{url}
\usepackage[hidelinks]{hyperref}
\usepackage[utf8]{inputenc}
\usepackage[small]{caption}
\usepackage{subfigure}
\usepackage{graphicx}
\usepackage{amsmath}
\usepackage{amsfonts}
\usepackage{amsthm}
\usepackage{booktabs}
\usepackage{algorithm}
\usepackage{algorithmic}
\usepackage{bookmark}
\usepackage{multirow}
\usepackage[switch]{lineno}
\usepackage[table]{xcolor}

\let\originalleft\left
\let\originalright\right
\renewcommand{\left}{\mathopen{}\mathclose\bgroup\originalleft}
\renewcommand{\right}{\aftergroup\egroup\originalright}

\newcommand{\expo}[1]{\exp\left(#1\right)}

\newcommand{\modeltheta}{\mathrm{\Theta}}
\newcommand{\absrp}{\sigma}

\newcommand{\numsamples}{N}

\newcommand{\timenear}{t_n}
\newcommand{\timefar}{t_f}
\newcommand{\deltatime}{\delta}

\newcommand{\ray}{\mathbf{r}}

\newcommand{\Ctrue}{C(\ray)}

\newcommand{\bc}{\mathbf{c}}
\newcommand{\bd}{\mathbf{d}}

\newcommand{\bo}{\mathbf{o}}

\newcommand{\bv}{\mathbf{v}}\newcommand{\bV}{\mathbf{V}}

\newcommand{\bx}{\mathbf{x}}

\newcommand{\nB}{\mathbb{B}}

\newcommand{\nR}{\mathbb{R}}

\newcommand{\cL}{\mathcal{L}}

\newcommand{\cR}{\mathcal{R}}

\newcommand{\figref}[1]{Fig.~\ref{#1}}
\newcommand{\secref}[1]{Sec.~\ref{#1}}

\newcommand{\eqnref}[1]{Equation~\eqref{#1}}
\newcommand{\tabref}[1]{Tab.~\ref{#1}}

\makeatletter
\DeclareRobustCommand\onedot{\futurelet\@let@token\@onedot}
\def\@onedot{\ifx\@let@token.\else.\null\fi\xspace}

\makeatother

\definecolor{orangeview}{RGB}{255,189,141}
\definecolor{blueview}{RGB}{143,170,220}
\definecolor{redvoxel}{RGB}{255,0,0}
\definecolor{purplevoxel}{RGB}{112,48,160}
\title{VGOS: Voxel Grid Optimization for View Synthesis from Sparse Inputs }

\author{
Jiakai Sun
\and
Zhanjie Zhang\and
Jiafu Chen\and
Guangyuan Li\and
Boyan Ji\and
Lei Zhao\footnote{Corresponding Authors}\and
Wei Xing$^{\ast}$\and
Huaizhong Lin
\affiliations
$^1$Zhejiang University\\
\emails
\{csjk, cszzj, chenjiafu, ji\_by, cszhl, wxing, linhz\}@zju.edu.cn,
lgy1428275037@163.com,
}
\begin{document}

\maketitle

\begin{abstract}
Neural Radiance Fields (NeRF) has shown great success in novel view synthesis due to its state-of-the-art quality and flexibility. However, NeRF requires dense input views (tens to hundreds) and a long training time (hours to days) for a single scene to generate high-fidelity images. Although using the voxel grids to represent the radiance field can significantly accelerate the optimization process, we observe that for sparse inputs, the voxel grids are more prone to overfitting to the training views and will have holes and floaters, which leads to artifacts. In this paper, we propose VGOS, an approach for fast (3-5 minutes) radiance field reconstruction from sparse inputs (3-10 views) to address these issues. To improve the performance of voxel-based radiance field in sparse input scenarios, we propose two methods: (a) We introduce an incremental voxel training strategy, which prevents overfitting by suppressing the optimization of peripheral voxels in the early stage of reconstruction. (b) We use several regularization techniques to smooth the voxels, which avoids degenerate solutions. Experiments demonstrate that VGOS achieves state-of-the-art performance for sparse inputs with super-fast convergence.
Code will be available at \url{https://github.com/SJoJoK/VGOS}.
\end{abstract}

\section{Introduction}
In 3D vision, novel view synthesis is a long-standing task that aims to synthesize a target image with an arbitrary target camera pose from given source images and their camera poses. Recently, Neural Radiance Fields (NeRF)~\cite{MildenhallSTBRN20}, a learning-based neural implicit representation, have emerged as a powerful tool yielding high-fidelity results on this task. However, NeRF requires tens to hundreds of dense inputs and hours to days of training time to get high-quality results. When considering real-world applications such as autonomous driving, AR/VR, and robotics that lack dense data and require real-time performance, NeRF's limitations of relying on dense input views and lengthy optimization time are even more magnified.

\begin{figure}[t]
	\centering
	\includegraphics[width=1\columnwidth]{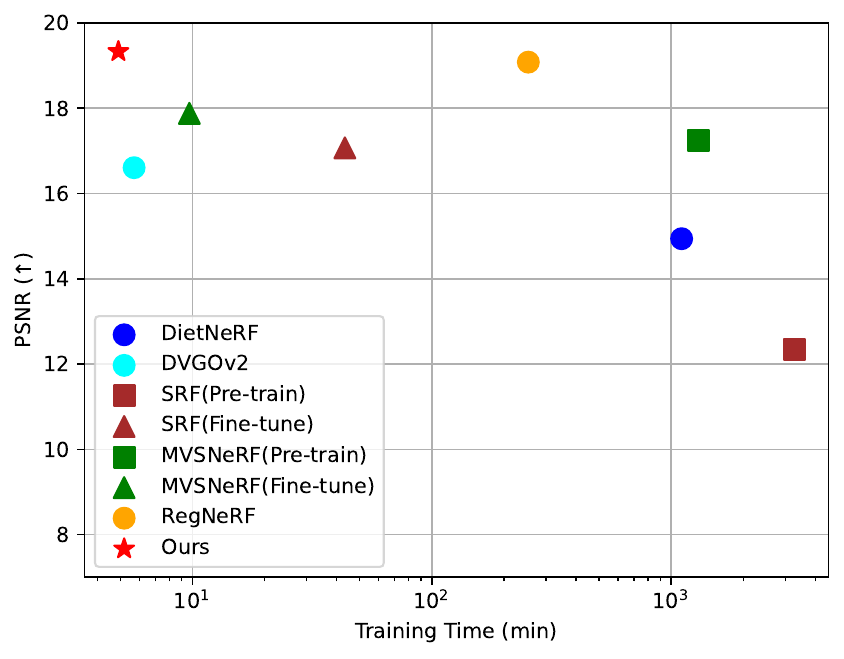} % Reduce the figure size so that it is slightly narrower than the column.
	\caption{The comparison between VGOS and previous methods on the LLFF dataset in 3-view settings. Note: For a fair comparison, the training time of each method is measured on our machine with a single NVIDIA RTX 3090 GPU using respective official implementations. Our model outperforms previous methods both in reconstruction speed (training time) and quality of results (PSNR) for sparse inputs.}
	\label{fig:comparison}
\end{figure}
\begin{figure*}[t]
	\centering
	\includegraphics[width=2\columnwidth]{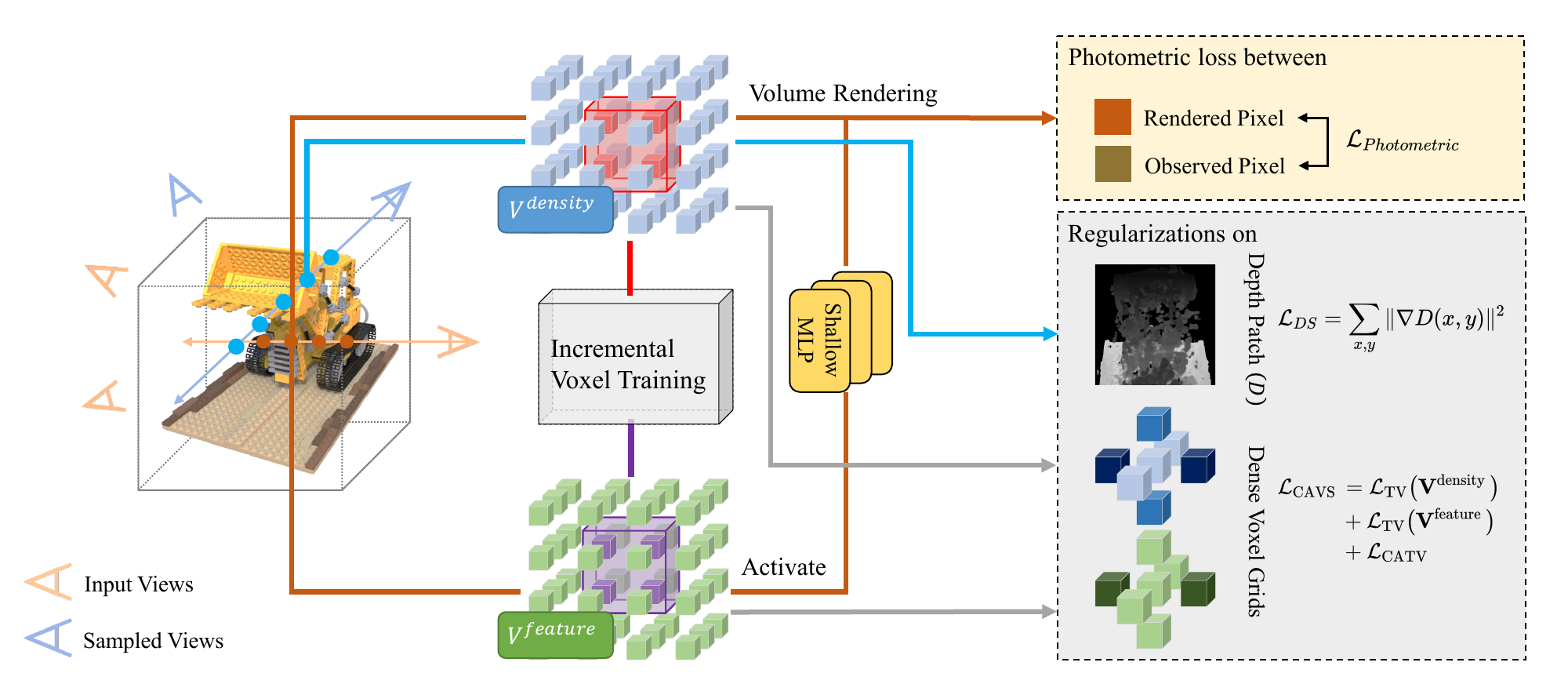} % Reduce the figure size so that it is slightly narrower than the column.
	\caption{Overview of VGOS architecture. Except for the photometric loss (\eqnref{equ:MSE}) from a given set of input images ({\color{orangeview}orange} views), the depth smoothness loss (\eqnref{equ:DS}) is imposed on the rendered depth patches (\eqnref{equ:depth_approxiamte}) from sampled views ({\color{blueview}blue} views), and the voxel girds are regularized by the proposed color-aware voxel smoothness loss (\eqnref{equ:CAVS}). Moreover, An incremental voxel training strategy is utilized to prevent overfitting by expanding the range of optimized voxels ({\color{redvoxel}red} and {\color{purplevoxel}purple} voxels) incrementally.}
	\label{fig:model}
\end{figure*}

To speed up the optimization, recent works~\cite{SunSC22,Chen2022ECCV,mueller2022instant,YuFTCR22} utilize explicit data structures to represent the radiance field, reducing the training time to minutes. However, these data structures designed to shorten the optimization process of the radiance field do not consider the performance for sparse inputs and still require dense inputs to obtain high-quality results.

To improve NeRF's performance on sparse inputs, several works~\cite{ChenXZZXYS21,YuYTK21,liu2022neuray} first pre-train a model on the multi-view images dataset of many scenes and then use the pre-trained model and the optional per-scene fine-tuning process to synthesize novel views for sparse inputs. Although these works have obtained promising results, acquiring pre-training data may be expensive, and the pre-training time is also very long. In addition, these methods may not generalize well for domains not covered by the pre-training data. Other works~\cite{Jain_2021_ICCV,Niemeyer2021Regnerf,kangle2021dsnerf,xu2022sinnerf} train the model from scratch for every new scene. To enhance the performance for sparse inputs, some works~\cite{Jain_2021_ICCV,Niemeyer2021Regnerf} regularize appearance or semantics by introducing models pre-trained on large-scale image datasets. Although these methods can generate high-quality rendering results, their results suffer from incorrect geometry, and the pre-trained model increases the method's complexity. Besides, some works leverage depth maps to supervise the optimization~\cite{kangle2021dsnerf} or augment training images~\cite{xu2022sinnerf}. The addition of depth information helps the model obtain relatively correct geometry from sparse inputs, but depth maps are not as easy to obtain as RGB images.

To overcome the aforementioned shortcomings and limitations,  we present an approach for fast radiance field reconstruction from sparse inputs,  namely VGOS. As shown in the \figref{fig:comparison}, our model achieves on-par high-quality results after minutes of training time compared with the previous state-of-the-art approaches, which take hours of per-scene optimization or days of generalizable pre-training. Specifically,  we directly optimize the voxel grids representing the radiance field~\cite{SunSC22}. However, for sparse inputs, the reconstruction of the radiance field (a) is more prone to overfitting to the training views, and (b) the voxel grids will have holes and floaters. In order to solve these two problems,  we propose two methods: (a) incremental voxel training strategy and (b) voxel smoothing method. With the improvement of the training strategy and the new regularization method,  our model achieves state-of-the-art performance for sparse input without any pre-trained model and with only RGB images as input.

Specifically, the incremental voxel training strategy is to freeze the optimization of peripheral voxels at the early stage of training and gradually thaw the peripheral voxels as the training progresses. This strategy prevents the voxels close to the cameras' near planes from overfitting to the training views, thus boosting the quality of radiance field reconstruction. The voxel smoothing method helps prevent degenerate solutions by regularizing the depth maps rendered from unobserved viewpoints~\cite{Niemeyer2021Regnerf} and penalizing the sharpness inside the voxel grids with the proposed color-aware voxel smoothness loss.

In summary, the main contributions of our work can be summarized as follows:
\begin{itemize}
    \item We propose an incremental voxel training strategy to prevent the voxels from overfitting to the training views by suppressing the optimization of peripheral voxels in the early stage of radiance field reconstruction.
    \item We propose a voxel smoothing method to avoid incorrect geometry by regularizing the dense voxel grids and utilizing depth smooth loss , which eliminates holes and floaters in the voxel grids, thus improving the quality of radiance field reconstruction in sparse input scenarios.
    \item Extensive experiments on different datasets demonstrate that our proposed model, even without any pre-trained model and extra inputs, achieves one to two orders of magnitude speedup compared to state-of-the-art approaches with on-par novel view synthesis quality.
\end{itemize}

\section{Related Work}
\subsection{Novel View Synthesis}
Novel view synthesis is a time-honored problem at the intersection of computer graphics and computer vision. Previous works use light field~\cite{LevoyH96,ShiHDKD14} and lumigraph~\cite{GortlerGSC96,BuehlerBMGC01} to synthesize novel views by interpolating the input images. Moreover, explicit representations, such as meshes~\cite{DebevecTM96,hu2021worldsheet}, voxels~\cite{sitzmann2019deepvoxels}, and multiplane images~\cite{mildenhall2019local,JohnFlynn2019DeepViewVS}, are introduced into this task. Recently, several works~\cite{sitzmann2019scene,niemeyer2020differentiable,yariv2020multiview,MildenhallSTBRN20} have introduced implicit representation and corresponding differentiable rendering methods due to their convenient end-to-end optimization and high-quality results. Among these works, Neural Radiance Fields~\cite{MildenhallSTBRN20} (NeRF) achieve photo-realistic rendering results by representing the radiance field as a multi-layer perceptron (MLP) and differentiable volume rendering method. Subsequent works have improved the performance of NeRF in many aspects, such as training on multi-resolution images~\cite{barron2021mip}, unconstrained images~\cite{martinbrualla2020nerfw,chen2022hallucinated}, unbounded scenes~\cite{barron2022mip}, dark scenes~\cite{mildenhall2022nerf}, and deforming scenes~\cite{park2021nerfies,park2021hypernerf}. However, NeRF and these variants require dense inputs to generate high-quality results, which is not always available in real-world applications.

\subsection{Fast Radiance Field Reconstruction}
Although NeRF can achieve high-fidelity rendering, it takes hours to days of training to reconstruct the radiance field for new scenes. Several works~\cite{SunSC22,YuFTCR22,Chen2022ECCV,mueller2022instant} use explicit or hybrid radiance field representations to reduce training time to a few minutes. DVGO~\cite{SunSC22} uses dense voxel grids and a shallow MLP to represent the radiance field, while DVGOv2~\cite{sun2022improved} re-implements some operations in CUDA to achieve improved performance. Plenoxels~\cite{YuFTCR22} uses a sparse voxel grid and coefficients of spherical harmonic for view-dependent colors to realize a fully explicit representation. TensoRF~\cite{Chen2022ECCV} achieves efficient radiance field reconstruction by decomposing the volume field and modeling the low-rank components. Instant-NGP~\cite{mueller2022instant} represents the radiance field as a multiresolution hash table and small neural networks, achieving convincing acceleration using C/C++ and fully-fused CUDA kernels. However, these acceleration methods do not reduce the dependence of radiance field reconstruction on dense inputs, while our approach performs high-quality novel view synthesis from sparse inputs in minutes of optimization.

\subsection{Sparse Input Radiance Field Reconstruction}
Many methods have been proposed to overcome the NeRF's dependence on dense inputs. Several works~\cite{YuYTK21,ChenXZZXYS21,chibane2021stereo} compensate for information scarcity from sparse inputs by pre-train a conditional model of the radiance field. PixelNeRF~\cite{YuYTK21} and SRF~\cite{chibane2021stereo} train convolutional neural network (CNN) encoders to extract features of the input images. MVSNeRF~\cite{ChenXZZXYS21} uses a 2D CNN to get 2D image features from the input images and then uses plane sweeping to obtain a cost volume which will be processed by a 3D CNN. These methods get promising results, but pre-training on multi-view image datasets is expensive and time-consuming. Besides, most of these methods require fine-tuning on new scenes, and the performance of these methods will decline when the data domain changes at test time.

On the other hand, a line of works~\cite{Jain_2021_ICCV,Niemeyer2021Regnerf,kangle2021dsnerf,xu2022sinnerf} use models pre-trained on large-scale image datasets and depth maps to train the radiance field from scratch. DietNeRF~\cite{Jain_2021_ICCV} uses prior knowledge about scene semantics learned by pre-trained CLIP ViT~\cite{radford2021learning} to constrain a 3D representation. RegNeRf~\cite{Niemeyer2021Regnerf} uses pre-trained Real-NVP~\cite{dinh2016density} to regularize the colors predicted at unseen viewpoints. DS-NeRF~\cite{kangle2021dsnerf} takes depth maps as input to supervise the reconstruction of the radiance field. Besides, SinNeRF~\cite{xu2022sinnerf} uses global structure prior provided by pre-trained DINO-ViT~\cite{caron2021emerging} and augments data using depth maps.

In addition, InfoNeRF~\cite{kim2022infonerf} is a prior-free model without any extra inputs, which regularizes the reconstruction of the radiance field by minimizing ray entropy and reducing information gain. However, this scheme requires the weights of all sampled points on rays. Therefore, reducing the number of sampling points is difficult, which is commonly used in NeRF acceleration approaches.

In contrast, our approach is 10$\times$-100$\times$ faster than state-of-the-art approaches with comparable high-quality results without expensive and time-consuming pre-train process and without additional input or pre-trained model to increase complexity. 
\section{Method}
Our approach, which builds upon DVGOv2~\cite{sun2022improved} (\secref{subsec:DVGO}), performs fast radiance field reconstruction from sparse RGB input images without any pre-trained model. We find that unexpected overfitting and holes and floaters of the voxel grids lead to degenerate solutions for sparse inputs. To prevent the radiance field from overfitting to the input views, we introduce an incremental voxel training strategy (\secref{subsec:incremental}) that suppresses the optimization of peripheral voxels. Moreover, we smooth the voxels (\secref{subsec:voxelSmooth}) by regularizing the predicted geometry from sampled views and the shape of the explicit radiance field. We depict an overview of our approach in \figref{fig:model}.
\subsection{Background}
\subsubsection{Neural Radiance Fields}
\label{subsec:NeRF}
A radiance field is a function that maps a 3D position $\bx$ and a viewing direction $\bd$ to the corresponding view-dependent emission color $\bc$ and volume density $\sigma$. NeRF~\cite{MildenhallSTBRN20} uses MLP to parameterize this function:
\begin{equation}
    \operatorname{MLP}_{\modeltheta} : (\mathbf{x}, \mathbf{d}) \to (\mathbf{c},\sigma),
\end{equation}
where $\modeltheta$ is the learnable MLP parameters. Note that the positional encoding~\cite{TancikSMFRSRBN20} is applied to $x$ and $d$ before the MLP to enable the MLP to represent higher frequency functions.

To synthesize novel views, NeRF uses volume rendering techniques. To be specific, the rendered color $\Ctrue$ of a target pixel is obtained by integrating colors and densities between near and far bounds $\timenear$ and $\timefar$ along a ray $\ray(t) = \bo + t\bd$ from the camera center $\bo$ through the pixel along direction $\bd$:
\begin{equation}
	\begin{aligned}
		\Ctrue = \int_{\timenear}^{\timefar}T(t)\absrp(\ray(t))\mathbf{c}(\ray(t),\mathbf{d})dt,
	\end{aligned}
\end{equation}
where $T(t)= \expo{-\int_{\timenear}^{t}\absrp(\ray(s))ds}$ is the accumulated transmittance along the ray from $\timenear$ to $t,$
and $\sigma(\cdot)$ and $\bc(\cdot, \cdot)$ indicate the density and color prediction of the radiance field $F_{\modeltheta}$, respectively.

In practice, the integral is approximated by quadrature:
\begin{equation}
	\begin{aligned}
        \label{equ:approxiamte_color}
		\hat{C}(\ray)=\sum_{i=1}^{\numsamples}T_i (1-\expo{-\absrp_i \deltatime_i}) \mathbf{c}_i,
	\end{aligned}
\end{equation}
where $T_i=\expo{- \sum_{j=1}^{i-1} \absrp_j \deltatime_j},$ $\numsamples$ is the number of samples points along the ray $\ray$;
$\absrp_i$, $\bc_i$ are the density and color of the $i^{th}$ sampled point,
and $\deltatime_i = t_{i+1} - t_i $ is the distance between adjacent samples.

NeRF's MLP can be optimized over a set of input images and their camera poses by minimizing the photometric MSE between the ground truth pixel color ${C}_{GT}(\ray)$ and the rendered color $\hat{C}(\ray)$:
\begin{equation}
\label{equ:MSE}
    \cL_{\text{Photometric}} = \frac{1}{|\cR|}\sum_{\ray \in \cR} \left\| \hat{C}(\ray) - {C}_{GT}(\ray)  \right\|^2\,,
\end{equation}
where $\cR$ denotes a set of rays.
\subsubsection{Direct Voxel Grid Optimization}
\label{subsec:DVGO}
It is time-consuming to query the color and density of each sampled point through MLP, so DVGO~\cite{SunSC22} is proposed to accelerate this process by representing the radiance field as voxel grids. Such an explicit scene representation is efficient to query color $\bc$ and density $\sigma$ for any 3D position $\bx$ with trilinear interpolation:
\begin{equation}
    \begin{aligned}
        \ddot{\sigma} &= \operatorname{interp}\left(\bx, \bV^{\text{density}}\right)~, \\
        \bc &= \operatorname{interp}\left(\bx, \bV^{\text{rgb}}\right)~, \\
        \sigma &= \log(1+\exp(\ddot{\sigma}+b))~,
    \end{aligned}
\end{equation}
where the shift $b = \log\left({\left(1 - \alpha_{\text{init}}\right)^{-\frac{1}{s}}} - 1\right)$ is the bias term determined by hyperparameter $\alpha_{\text{init}}$ and voxel size $s$,
$\bV^{\text{density}}$ and $\bV^{\text{rgb}}$ are the voxel grids storing raw density $\ddot{\sigma}$ before applying the density activation and color, respectively.

In practice, DVGO uses a coarse-to-fine training strategy. In the fine stage, a shallow MLP is used to process viewing-direction $\bd$ and feature $f$ from a feature voxel grid $\bV^{\text{feature}}$ to model view-dependent color emission.
\begin{equation}
    \bc = \operatorname{MLP}_{\Theta}\left(\operatorname{interp}(\bx, \bV^{\text{feature}}), \bx, \bd\right) ~,
\end{equation}
where $\Theta$ is the learnable MLP parameters.

Subsequent work, namely DVGOv2~\cite{sun2022improved}, improve DVGO by re-implementing part of the Pytorch operations with CUDA and extending it to support forward-facing and unbounded inward-facing capturing.
\begin{figure}[t]
	\centering
	\includegraphics[width=1\columnwidth]{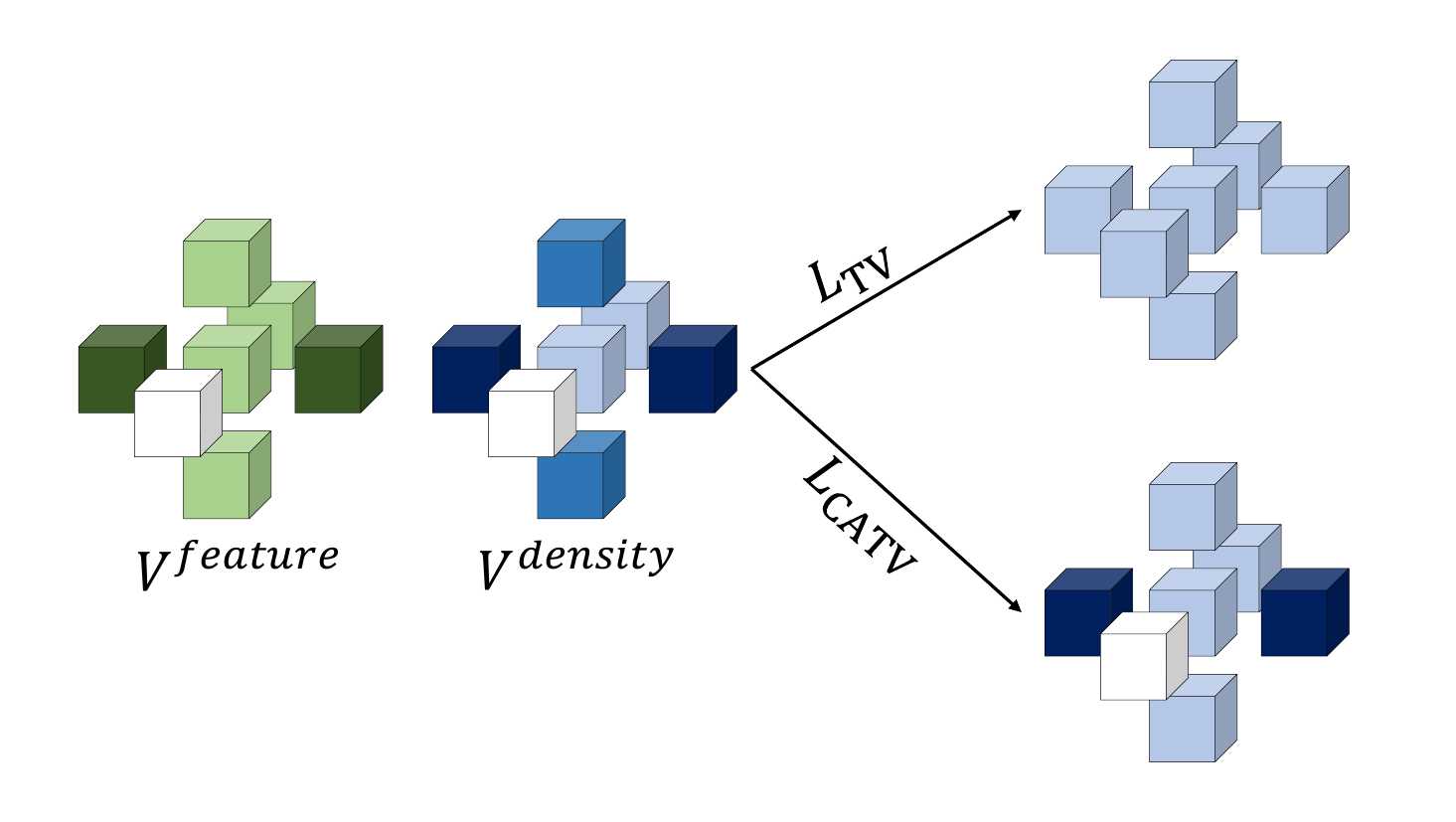} 
	\caption{
    Visual comparisons between $\cL_{\text{TV}}$ and $\cL_{\text{CATV}}$ regularization.
    $\cL_{\text{TV}}$ doesn't utilize the information of $\bV^{\text{feature}}~,$
    while $\cL_{\text{CATV}}$ takes advantage of the correlation between $\bV^{\text{density}}$ and $\bV^{\text{feature}}~.$}
	\label{fig:CATV}
\end{figure}

\begin{figure*}[t]
	\centering
	\includegraphics[width=2\columnwidth]{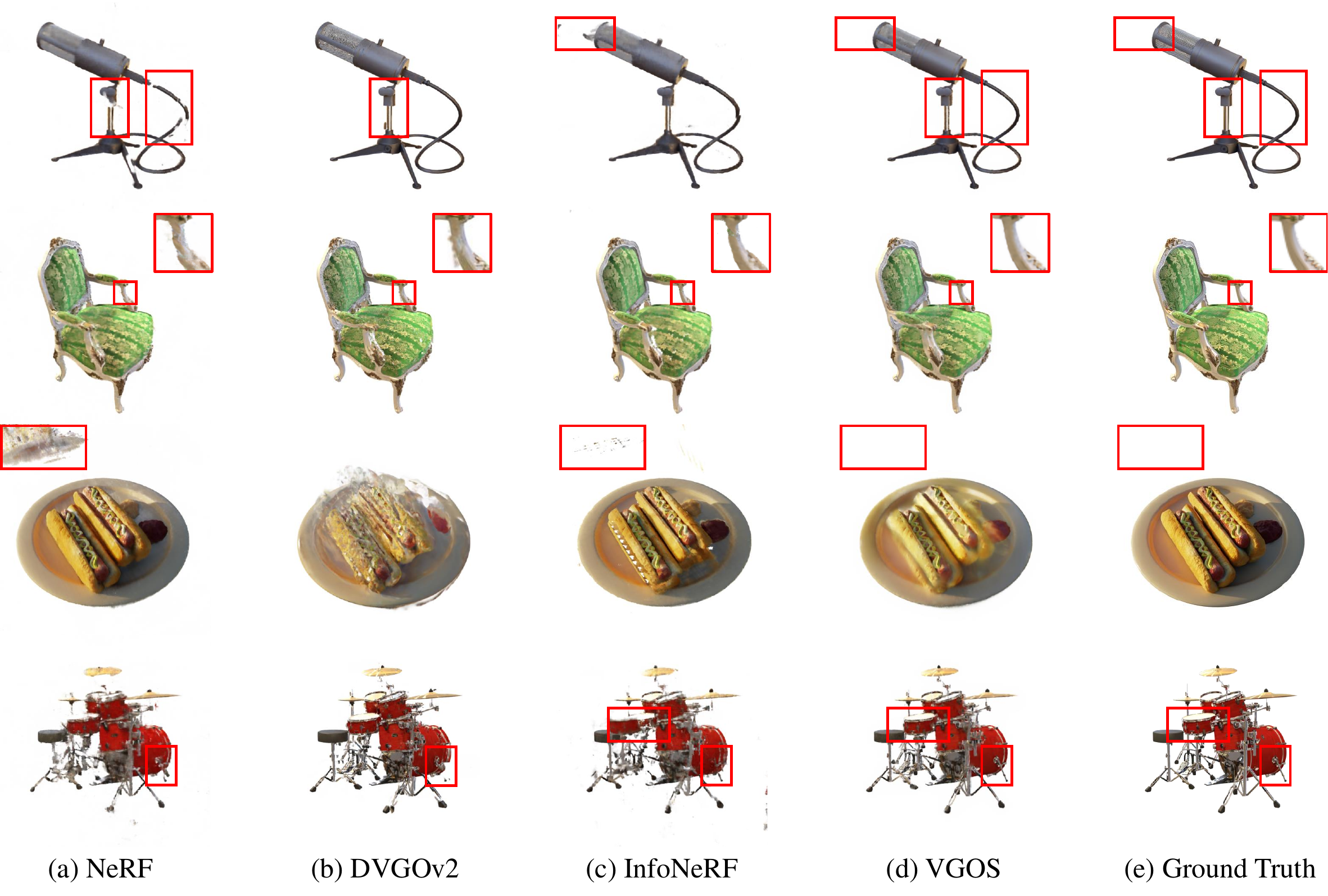}
	\caption{Qualitative comparison on Realistic Synthetic 360° in the 4-view setting. All the experiments are performed with the same inputs. Please zoom in for more details.}
	\label{fig:blender_cpr}
\end{figure*}

\subsection{Incremental Voxel Training}
\label{subsec:incremental}
Although DVGO uses various techniques to avoid degenerate solutions, the radiance field will overfit to input views for sparse scenarios. Specifically, we find that for sparse inputs, the peripheral voxels close to the camera near planes have high density values at the initial stage of training to reproduce the input views. However, the high density value of the outer voxels hinders the optimization of the inner voxels, which makes it difficult for the radiance field to converge to the correct geometry so that the quality of rendering results at novel views will decline.

We propose a simple yet non-trivial incremental voxel training strategy to solve the above-mentioned problem. For the voxel grids $\bV \in \nR^{C \times N_x \times N_y \times N_z}$ representing the radiance field, where $C$ is the dimension of the modality, $N_x\cdot N_y \cdot N_z$ is the total number of voxels, we define an expanding bounding box $\nB$ whose corner points are $P_{min}, P_{max}\in \nR^3$:
\begin{equation}
    \begin{aligned}
        P_{min}&=(P_{min\_init}\times(1-r(i)))\odot (N_x, N_y, N_z)~,\\
        P_{max}&=(P_{max\_init}\times(1-r(i))+r(i))\\
               &\odot (N_x, N_y, N_z)~,
    \end{aligned}
\end{equation}
where $P_{min\_init}\in[0,1]^3$ and $P_{max\_init}\in[0,1]^3$ are the initial ratio of the expanding bounding box $\nB$,
and $r(i)=min(\frac{i}{M},1)$ determine the range of the bounding box $\nB$,
where $i$ is the current training iteration and $M$ is the pre-defined max iteration steps of the increment process.
We only optimize the voxels inside the bounding box $\nB$; this training strategy freezes the optimization of the peripheral voxels in the early training, avoiding overfitting and leading to better rendering results at novel views.
We set $M=256$ for all scenes,
$P_{min\_init}=(0.2,0.2,0.2), P_{max\_init}=(0.8,0.8,0.8)$ for bounded inward-facing scenes and 
$P_{min\_init}=(0,0,0.995), P_{max\_init}=(1,1,1)$ and $M=256$ for forward-facing scenes in our experiments.
\subsection{Voxel Smoothing}
Although we use the incremental voxel training strategy to alleviate the overfitting, if we only use the photometric MSE loss (\eqnref{equ:MSE}) to supervise the training from sparse inputs, the radiance field will still overfit to the input views. To solve this problem, we propose a novel color-aware voxel smoothness loss on the dense voxel grids and utilize the depth smoothness loss on the sampled views to smooth the voxels.
\subsubsection{Regularization on Dense Voxels}
\label{subsec:voxelSmooth}
To prevent the outliers and noises in the explicit model, previous works~\cite{YuFTCR22,Chen2022ECCV,sun2022improved} utilize total variation (TV) loss~\cite{rudin1994total}:
\begin{equation}
    \cL_{\text{TV}}(\bV)=\sum_{\substack{\bv \in \bV}} \Delta(\bv)~,
\end{equation}
with $\Delta(\bv)$ shorthand for the mean of the loss (L1, L2, or Huber Loss) between the value in voxel $\bv$ and its six nearest-neighbor voxels and $\bV$ indicates the voxel grids storing density, color or feature, which is indeed effective.
However, these works calculate the TV loss of density and color separately, not taking advantage of the correlation between density and color in the explicit radiance field.

We observe that, in the radiance field, the density change is not smooth where the color changes sharply. According to the above observation, we propose color-aware total variance (CATV) loss, which uses the activated value in the color voxel grid to guide TV loss of the density voxel grid and is formulated as:
\begin{equation}
    \begin{aligned}
        \label{equ:CATV}
        F_{\text{CA}}(\bV,\bv)&=\Delta_{\text{activate}}(\bv),\bv \in \bV~,\\
        \cL_{\text{CATV}}&=\sum_{\substack{\bv \in \bV^{\text{density}}}}e^{-F_{\text{CA}}(\bV^{\text{feature}},\bv)}\Delta(\bv)~,
    \end{aligned}				
\end{equation}
with $\Delta_{\text{activate}}(\bv)$ indicates that the activated values are used calculating $\Delta(\bv)$.
In practice, we use L1 loss in $F_{\text{CA}}$, and Huber Loss is in $\cL_{\text{CATV}}$. Sigmoid is used in $F_{\text{CA}}$ to normalize the feature values to $[0,1]$ and align the choices of activation functions in DVGO. In \figref{fig:CATV}, we show the differences between using $\cL_{\text{TV}}$ and $\cL_{\text{CATV}}$ to regularize $\bV^{\text{density}}$.

To ensure flexibility, we use both $\cL_{\text{TV}}$ and $\cL_{\text{CATV}}$.
Therefore, the color-aware voxel smoothness (CAVS) loss which is used to regularize the dense voxels is formulated as:
\begin{equation}
    \begin{aligned}
    \label{equ:CAVS}
    \cL_{\text{CAVS}}&=\lambda_{\text{TVF}}\cL_{\text{TV}}(\bV^{\text{feature}})+
                            \lambda_{\text{TVD}}\cL_{\text{TV}}(\bV^{\text{density}})\\
                            &+\lambda_{\text{CATV}}\cL_{\text{CATV}}~,
    \end{aligned}                       
\end{equation}
where $\lambda_{\text{TVF}}$, $\lambda_{\text{TVD}}$ and $\lambda_{\text{CATV}}$ are the corresponding weights.
Since computing $\cL_{\text{CAVS}}$ is time-consuming, we implement it in CUDA kernel to speedup the process. Besides, we only backpropagate the gradient of $\cL_{\text{CATV}}$ to $\bV^{\text{density}}$.
We set $\lambda_{\text{TVD}}=5\cdot10^{-4}~, \lambda_{\text{TVF}}=\lambda_{\text{CATV}}=5\cdot10^{-5}$ in coarse-stage training 
and $\lambda_{\text{TVD}}=5\cdot10^{-5}~, \lambda_{\text{TVF}}=10^{-5},\lambda_{\text{CATV}}=5\cdot10^{-6}$ in fine-stage training for bounded inward-facing scenes.
For forward-facing scenes which only need fine-stage training, we set $\lambda_{\text{TVD}}=5\cdot10^{-5}~,$and $\lambda_{\text{TVF}}=\lambda_{\text{CATV}}=5\cdot10^{-6}~.$
\subsubsection{Regularization on Sampled Viewpoints}
\label{subsec:depthSmooth}
The piecewise-smooth of geometry is a classic hypothesis in depth and disparity estimation~\cite{scharstein2002taxonomy}. Hence we utilize the depth smoothness (DS) loss introduced by RegNeRF~\cite{Niemeyer2021Regnerf} on the unseen views to improve scene geometry.

To get unobserved views, we sample camera pose $p$ $\sim$ $\pi$ where $\pi$ is the distribution of camera poses if $\pi$ is available.
For bounded inward-facing scenes such as one from the Realistic Synthetic 360$^\circ$ dataset~\cite{MildenhallSTBRN20},
$\pi$ is the uniform distribution over the hemisphere with the known radius.
For forward-facing scenes like those from the LLFF dataset~\cite{mildenhall2019local},
$\pi$ is the uniform distribution over a 2D plane with given boundaries.
If $\pi$ is not available, we generate new poses by interpolating between input poses.

We can estimate depth $\hat{d}$ along the ray $\ray$ cast from the sampled camera pose the similar way we render color in \eqnref{equ:approxiamte_color}:
\begin{equation}
    \label{equ:depth_approxiamte}
		\hat{d}(\ray)=\sum_{i=1}^{\numsamples}T_i (1-\expo{-\absrp_i \deltatime_i})~.
\end{equation}
By estimating depth from sets of neighboring rays, we can render depth patches and regularize them by the DS loss:
\begin{equation}
    \label{equ:DS}
    \cL_{\text {DS}}=\frac{\lambda_{\text{DS}}}{|\cR|}\sum_{\ray_c \in \cR}\sum_{(x, y)}\|\nabla D(x, y)\|^{2}~,
\end{equation}
where $\cR$ indicates a set of rays cast from the sampled poses, $D$ is the depth patch centered at $\ray_c$, and $\lambda_{\text{DS}}$ is the loss weight.
In practice, finite difference formula is used to compute $\nabla D$.
We set $\lambda_{\text{DS}}=5\cdot10^{-4}$ in coarse-stage training 
and $\lambda_{\text{DS}}=10^{-5}$ in fine-stage training for bounded inward-facing scenes.
For forward-facing scenes, we set $\lambda_{\text{DS}}=5\cdot10^{-4}$.
\subsection{Total Loss Function}
The total loss function of our model is given by:
\begin{equation}
    \cL_{\text {Total}}=\cL_{\text{Photometric}}+\cL_{\text{CAVS}}+\cL_{\text {DS}}~,
\end{equation}
note that the hyperparameters to balance the regularization terms have been included in \eqnref{equ:CAVS} and \eqnref{equ:DS}.
Besides, $\cL_{\text{Photometric}}$ uses the rays from the input views, and $\cL_{\text{CAVS}}$ utilizes the rays from the sampled views.

\section{Experiments}
\subsection{Datasets and Evaluations}
We perform experiments on inward-facing scenes from the Realistic Synthetic 360$^\circ$ dataset~\cite{MildenhallSTBRN20} and forward-facing scenes from the LLFF dataset~\cite{mildenhall2019local}.
\paragraph{Realistic Synthetic 360$^\circ$ Dataset}
The Realistic Synthetic 360$^\circ$ dataset contains pathtraced images of 8 synthetic scenes with complicated geometry and realistic non-Lambertian materials. Each scene has 400 images rendered from inward-facing virtual cameras with different viewpoints. Following the protocol of InfoNeRF~\cite{kim2022infonerf}, we randomly sample 4 views out of 100 training images as sparse inputs and evaluate the model with 200 testing images.
\paragraph{LLFF Dataset}
The LLFF Dataset consists of 8 complex real-world scenes captured by a handheld cellphone. Each scene has 20 to 62 forward-facing images. We hold out 1/8 of the images as test sets following the standard protocol~\cite{MildenhallSTBRN20} and report results for 3 input views randomly sampled from the remaining images.
\paragraph{Metrics}
We measure the mean of peak signal-to-noise ratio (PSNR), structural similarity index measure (SSIM)~\cite{WangBSS04}, and learned perceptual image patch similarity (LPIPS)~\cite{ZhangIESW18} to evaluate our model.
\subsection{Implementation Details}
We implement our model on the top of DVGOv2 codebase using Pytorch~\cite{paszke2019pytorch}.
Following DVGO, We use the Adam~\cite{KingmaB14} to optimize the voxel grids with the initial learning rate of $0.1$ for all voxels and $10^{-3}$ for the shallow MLP and exponential learning rate decay is applied.

For scenes in the Realistic Synthetic 360$^\circ$ dataset, we train the voxel grids for 5K iterations with a batch size of $2^{13}$ rays for input views and $2^{14}$ rays for sampled views in both stages.

For scenes in the LLFF dataset, we train the voxel grids for 9K iterations with a batch size of $2^{12}$ rays for input views and $2^{14}$ rays for sampled views in only one stage.

Please refer to the supplementary material for more details.
\subsection{Comparisons}
Following InfoNeRF~\cite{kim2022infonerf}, the presented metrics for comparisons are the average score of five experiments with different viewpoint samples.
\subsubsection{Realistic Synthetic 360° Dataset}
\begin{table}[t]
    \centering
        \scalebox{0.85}{
            \begin{tabular}{lcccc}
                \toprule
                Model &PSNR$\uparrow$ &SSIM$\uparrow$ &LPIPS$\downarrow$ &Training Time$\downarrow$\\
                \midrule
                NeRF        &15.93 &0.780 &0.320 &2 hrs$^\ast$\\
                DietNeRF    &16.06 &0.793 &0.306 &19 hrs\\
                PixelNeRF   &16.09 &0.738 &0.390 &3-4 days$^\star$+10 hrs\\
                DVGOv2      &17.19 &0.767 &\underline{0.223} &\underline{4 mins}\\
                InfoNeRF    &\underline{18.62} &\underline{0.811} &0.230 &4 hrs\\
                VGOS(ours)  &\textbf{18.91} &\textbf{0.825} &\textbf{0.205} &\textbf{3 mins}\\
                \bottomrule
            \end{tabular}}
    \caption{Quantitative comparison on Realistic Synthetic 360° in the 4-view setting. The asterisk ($\ast$) denotes that early-stopping is used instead of the default setting. The star ($\star$) denotes the generalizable pre-training time. Bold and underline indicate the best and the second-best values for each metric.}
    \label{tab:blender_comparsion}
\end{table}
We compare our model with NeRF~\cite{MildenhallSTBRN20},
DietNeRF~\cite{Jain_2021_ICCV},
PixelNeRF~\cite{YuYTK21},
InfoNeRF~\cite{kim2022infonerf},
and DVGOv2~\cite{SunSC22,sun2022improved} on the Realistic Synthetic 360° dataset in the 4-view setting.
Since PixelNeRF is pre-trained on the DTU dataset, we fine-tuned it for 20K iterations similar to ~\cite{kangle2021dsnerf} for improved performance.

\tabref{tab:blender_comparsion} presents the overall quantitative results,
and \figref{fig:blender_cpr} shows the qualitative results.
As the baseline, NeRF has degenerate solutions for sparse inputs. DietNeRF and PixelNeRF outperform the baseline relatively by introducing pre-trained models. Although DVGOv2 aims to accelerate the reconstruction process, it achieves superior results to NeRF in sparse input scenarios, which we observe as another advantage of explicit models. InfoNeRF outperforms the previous methods in terms of all image quality metrics. However, it takes twice the training time than NeRF and is unsuitable for common acceleration approaches since it requires the weights of all sampled points on rays. Our model achieves state-of-the-art performance with an outstanding convergence speed.
\subsubsection{LLFF Dataset}
\begin{table}[t]
    \centering
        \scalebox{0.85}{
            \begin{tabular}{lcccc}
                \toprule
                Model &PSNR$\uparrow$ &SSIM$\uparrow$ &LPIPS$\downarrow$ &Training Time$\downarrow$\\
                \midrule
                PixelNeRF   &16.17 &0.438 &0.512 &3-4 days$^\star$+10 hrs\\
                SRF         &17.07 &0.436 &0.529 &2-3 days$^\star$+43 mins\\
                MVSNeRF     &17.88 &0.584 &\textbf{0.327} &1-2 days$^\star$+10 mins\\
                Mip-NeRF       &14.62 &0.351 &0.495 &14 hrs\\
                DietNeRF       &14.94 &0.370 &0.496 &18 hrs\\
                DVGOv2         &16.60 &0.566 &0.422 &\underline{6 mins}\\
                RegNeRF        &\underline{19.08} &\underline{0.587} &\underline{0.336} &4 hrs\\
                VGOS(ours)     &\textbf{19.35} &\textbf{0.620} &0.432 &\textbf{5 mins}\\
                \bottomrule
            \end{tabular}}
    \caption{Quantitative comparison on LLFF in the 3-view setting. The star ($\star$) denotes the generalizable pre-training time. Bold and underline indicate the best and the second-best values for each metric. Note that the training time of Mip-NeRF can be reduced to 3 hours if early-stopping is applied, but we follow the baseline implement of RegNeRF's official code which trains Mip-NeRF for 250K iters with a batch size of $2^{12}$ rays.}
    \label{tab:llff_comparsion}
\end{table}

We compare our model with Mip-NeRF~\cite{barron2022mip},
DietNeRF~\cite{Jain_2021_ICCV},
PixelNeRF~\cite{YuYTK21},
SRF~\cite{chibane2021stereo},
MVSNeRF~\cite{ChenXZZXYS21},
RegNeRF~\cite{Niemeyer2021Regnerf} and DVGOv2~\cite{SunSC22,sun2022improved} on the LLFF Dataset in the 3-view setting.
Similar to the experiments on the Realistic Synthetic 360° Dataset, we fine-tune PixelNeRF, SRF, and MVSNeRF on each scene of the LLFF dataset to handle the domain shift issue since these methods are pre-trained on the DTU dataset.

The overall quantitative results are presented in \tabref{tab:llff_comparsion}. Besides, we provide the qualitative results in the supplementary material. Our model is superior to previous works in each metric except for LPIPS, which measures human perception. However, pre-trained model extracting high-level information is not used in our approach, which is a trade-off between complexity and performance, leading to relatively higher LPIPS on our model's evaluation results.
\subsection{Ablation Study}
\subsubsection{Effectiveness of Proposed Components}
\begin{table}[t]
    \centering
        \scalebox{0.85}{
        \begin{tabular}{lcccccc}
            \toprule
            Variant &  \textit{Inc.} &  $\cL_{\text {DS}}$ &  $\cL_{\text{CAVS}}$&PSNR$\uparrow$ &SSIM$\uparrow$ &LPIPS$\downarrow$ \\
            \midrule
            Baseline & $\times$ & $\times$ & $\times$ & 14.47 & 0.5684 & 0.5830 \\ 
            \textit{w/~Inc.} & $\checkmark$ & $\times$ & $\times$ & 15.86 & 0.6303 & 0.5281 \\ 
            \textit{w/~}$\cL_{\text {DS}}$ & $\times$ & $\checkmark$ & $\times$ & 16.91  & 0.6756 & 0.5104  \\ 
            \textit{w/~}$\cL_{\text{CAVS}}$ & $\times$ & $\times$ & $\checkmark$ & \underline{20.08}  & \underline{0.7939} & 0.4258 \\ 
            \textit{w/o~Inc.} & $\times$ & $\checkmark$ & $\checkmark$ & 18.57 & 0.7688 & 0.4404\\ 
            \textit{w/o~}$\cL_{\text {DS}}$ & $\checkmark$ & $\times$ & $\checkmark$ & 17.84   & 0.7357 & 0.4730 \\ 
            \textit{w/o~}$\cL_{\text{CAVS}}$ & $\checkmark$ & $\checkmark$ & $\times$ & 19.79 & 0.7663 & \underline{0.4183} \\ 
            Full Model& $\checkmark$ & $\checkmark$ & $\checkmark$ & \textbf{21.82} & \textbf{0.8220} & \textbf{0.3869}
            \\ \hline
        \end{tabular}
        }
    \caption{Ablation study on the \textit{room} scene in the 3-view setting.}
    \label{tab:room_ablation}
\end{table}
We conduct ablation studies on the \textit{room} scene to evaluate the contributions of each component of our proposed model.
\tabref{tab:room_ablation} report the results with various combinations of the incremental voxel training strategy (\textit{Inc.}), $\cL_{\text {DS}}$ and $\cL_{\text{CAVS}}$.
Besides, we provide more results and vision comparisons in the supplementary material.

With (PSNR+X) indicating the PSNR improvement compared to the baseline, we can observe the following:

\textit{w/~Inc.}~improve performance~(PSNR+1.39),~which demonstrates the effectiveness of~\textit{Inc.}.~Similarly,~\textit{w/~}$L_{CAVS}$ results in a substantial boost in performance (PSNR+5.61), highlighting the effectiveness of the $\cL_{\text{CAVS}}$ regularization term.

The effects of these three components are not strictly orthogonal. For instance, while combining $\cL_{\text {DS}}$ and \textit{Inc.} leads to improved performance (\textit{w/o~}$\cL_{\text{CAVS}}$ vs. \textit{w/~Inc.}), adding $\cL_{\text {DS}}$ to a model that only utilizes $\cL_{\text{CAVS}}$ results in a decrease in performance (\textit{w/o~Inc.} vs. \textit{w/~}$\cL_{\text{CAVS}}$)

The full model, which combines all three components, achieves the best performance overall. This indicates the effectiveness of our proposed model.
\subsubsection{Effectiveness of $\cL_{\text{CATV}}$}
\begin{figure}[t]
	\centering
	\includegraphics[width=1\columnwidth]{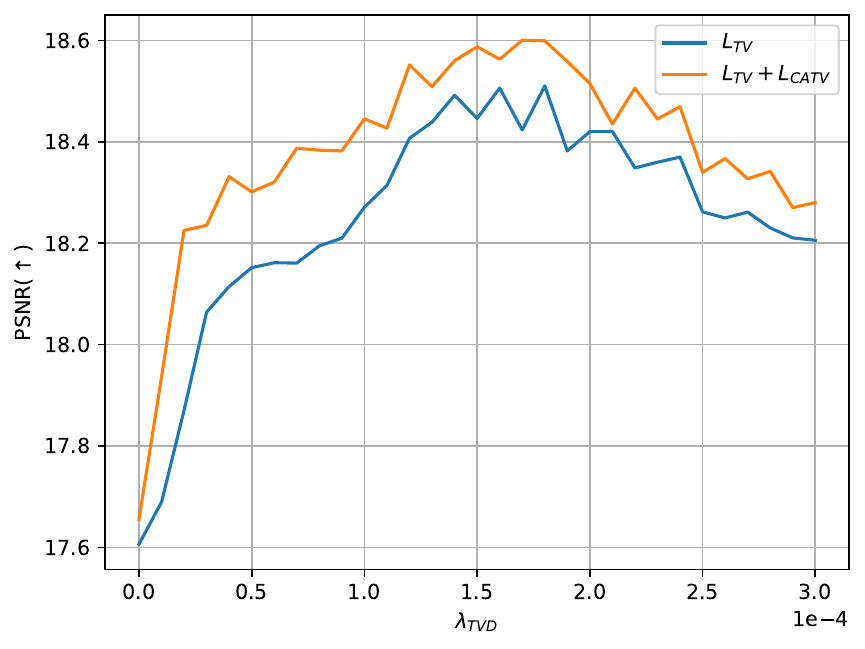}
	\caption{Comparison of $\cL_{\text{TV}}$ and $\cL_{\text{TV}}+\cL_{\text{CATV}}$ on the \textit{flower} scene in the 3-view setting, with $\lambda_{\text{CATV}}$ set to 0 and $5\cdot10^{-6}$ separately.}
	\label{fig:cavs_cpr}
\end{figure}

We conduct ablation studies on the \textit{flower} scene to evaluate the contribution of $\cL_{\text{CATV}}$.
As illustrated in the \figref{fig:cavs_cpr}, the performance of the model is improved by adding $\cL_{\text{CATV}}$, regardless of $\lambda_{\text{TVD}}$.
This demonstrate the effectiveness of the $\cL_{\text{CATV}}$ regularization term.
Note that the gradient of $\cL_{\text{CATV}}$ is only backpropagated to $\bV^{\text{density}}$ in our implementation (\eqnref{equ:CATV}), so $\lambda_{\text{TVF}}$ remains fixed at $5\cdot10^{-6}$.
\section{Conclusion}
NeRF suffers from a long training time and the requirement of dense inputs. To overcome the above shortages, we propose VGOS, an approach to improve the performance of the voxel-based radiance field from sparse inputs. By directly optimizing voxel grids, the incremental voxel training strategy, and the voxel smoothing method, VGOS is $10\times-100\times$ faster than previous few-shot view synthesis methods with state-of-the-art render quality while avoiding the degenerate solutions for explicit radiance field methods in sparse input scenarios.
\section*{Acknowledgements}
\bibliographystyle{named}
This work was supported in part by the projects No. 2020YFC1523202-2, 2021YFF0900604, 19ZDA197, LY21F020005, 2021009, 2019011,2022C01222, Research and Application Demonstration of Key Technologies in Digital Archaeology of Ningbo Prehistoric Sites, Key Technologies and Product Research and Development for Cultural Relics Protection and Trading Circulation, MOE Frontier Science Center for Brain Science \& Brain-Machine Integration (Zhejiang University), National Natural Science Foundation of China (62172365). 
\bibliography{neural_rendering}

\end{document}